\documentclass[a4paper,fleqn]{cas-dc}

\usepackage[authoryear]{natbib}
\usepackage{graphicx}
\usepackage{epstopdf}
\usepackage{subfig}    
\usepackage{float}  
\usepackage{tikz}
\usepackage{caption}
\usepackage{ragged2e} 

\def\tsc#1{\csdef{#1}{\textsc{\lowercase{#1}}\xspace}}
\tsc{WGM}
\tsc{QE}

\begin{document}
\let\WriteBookmarks\relax
\def\floatpagepagefraction{1}
\def\textpagefraction{.001}
\let\printorcid\relax 

\shorttitle{A method for detecting dead fish on large water surfaces based on improved YOLOv10}    


\title[mode = title]{A method for detecting dead fish on large water surfaces based on improved YOLOv10}

\author[1,2,3]{\textcolor{black}{Qingbin Tian}}[type=editor,
    orcid=0009-0004-1106-5560]
\ead{tianqingbin@cau.edu.cn} 
\credit{Conceptualization of this study, Methodology, Software}

\author[1,2,3]{\textcolor{black}{Yukang Huo}}[]

\author[1,2,3]{\textcolor{black}{Mingyuan Yao}}[]

\author[5]{\textcolor{black}{Yugang Cai}}

\author[1,2,3,4]{\textcolor{black}{Haihua Wang}\textsuperscript{*}}

\address[1]{National Innovation Center for Digital Fishery, China Agricultural University, Beijing 100083, China}
\address[2]{College of Information and Electrical Engineering, China Agricultural University, Beijing 100083, China}
\address[3]{Key Laboratory of Smart Farming Technologies for Aquatic Animal and Livestock, Ministry of Agriculture and Rural
	Affairs, Beijing, 100083, P.R, China}
\address[4]{Beijing Engineering and Technology Research Center for Internet of Things in Agriculture, Beijing, 100083, P.R, China}
\address[5]{Junshan Ecological Fishery Group Co., Ltd., Hunan, China}

\cortext[1]{Corresponding author} 
\cortext[2]{Principal corresponding author} 

\begin{abstract}
The presence of dead fish can lead to various issues such as water pollution and disease transmission, necessitating prompt detection and removal. Traditional methods for detecting dead fish are often limited by manpower and time, and struggle to effectively handle the complexities of aquatic environments. This paper proposes an end-to-end detection model based on an enhanced YOLOv10 framework, designed to rapidly and accurately detect dead fish across large water surfaces. Key enhancements include: (1) Replacing YOLOv10's backbone network with FasterNet to reduce model complexity while maintaining high detection accuracy; (2) Improving feature fusion in the Neck section through enhanced connectivity methods and replacing the original C2f module with CSPStage modules; (3) Adding a compact target detection head to enhance the detection performance of smaller objects. Experimental results demonstrate significant improvements in P(precision), R(recall), and AP(average precision) compared to the baseline model YOLOv10n. Furthermore, our model outperforms other models in the YOLO series by significantly reducing model size and parameter count, while sustaining high inference speed and achieving optimal AP performance. The model facilitates rapid and accurate detection of dead fish in large-scale aquaculture systems. Finally, via ablation experiments, we systematically analyze and assess the contribution of each model component to the overall system performance.
\end{abstract}




\maketitle

\section{Introduction}

Aquaculture is a rapidly growing global industry, with China being the largest producer and exporter of aquatic products \cite{LIU2017161}. Fish and fisheries play crucial roles in food security, societal well-being, and environmental health \cite{doi:10.1139/er-2015-0064}. As integral components of aquaculture, fish are essential for ensuring production safety and nutritional strategies. However, the industry's growth presents challenges, including environmental pollution and pathogen outbreaks that frequently lead to fish mortality. These issues not only impact aquatic ecosystems directly but also pose significant risks to surrounding environments and human health, thereby constraining the sustainable development of aquaculture.

Traditional methods for detecting dead fish typically involve manual observation, which is time-consuming, labor-intensive, and inefficient, leading to inherent uncertainties.Recently, there have been substantial advancements in image-based object detection due to rapid developments in deep learning \cite{krizhevsky2017imagenet}. Consequently, researchers have increasingly turned to deep learning approaches for dead fish detection:
Addressing the challenge of identifying dead fish in large-scale net cages, \cite{yu2020adaptive} introduced a technique utilizing SSD-MobileNet for the detection of dead fish on the water surface. This approach leverages hardware network architecture search (NAS) and NetAdapt's architecture to optimize network design through automated search algorithms. It demonstrates robust performance in both detection accuracy and speed.
Furthermore, \cite{zhao2022lightweight} developed a lightweight end-to-end model for dead fish detection using deep neural networks. By integrating deformable convolutions and enhancing YOLOv4, the model achieves significant reductions in network parameters and computational load. Experimental results underscore the model's high accuracy and effective real-time performance in underwater images.

The integration of drones with object detection technology for real-time surface monitoring offers a solution to swiftly identify and retrieve dead fish. This approach effectively mitigates water pollution from fish deaths, prevents large-scale fish mortality, and enhances economic benefits. To achieve such advanced object detection capabilities, significant developments in neural network architectures have been pivotal. For instance, \cite{girshick2014rich} introduced the Region Convolutional Neural Network (RCNN), which combines selective search with convolutional neural networks (CNN) for object detection. Building on this, \cite{he2015spatial} proposed the SPPNet algorithm, which applies a single convolution operation to the entire input image.This approach reduces redundant computations and significantly boosts detection speed compared to RCNN, while maintaining similar accuracy. Subsequently, Faster RCNN, introduced by \cite{ren2015faster}, utilizes a Region Proposal Network (RPN) to enhance the efficiency of candidate box generation, feature extraction, and bounding box regression within a unified framework, enabling end-to-end training and detection. Additionally, \cite{he2017mask} proposed Mask RCNN, which improves upon ROI Pooling with ROI Align and eliminates quantization operations, leading to substantial gains in detection accuracy. These algorithms are categorized as two-stage object detection methods, which typically exhibit lower detection speeds.

In contrast, \cite{redmon2016you} introduced YOLOv1, the first single-stage object detection algorithm, which performs object detection with just one pass through a neural network, ensuring high detection speed.Another single-stage algorithm, SSD, was proposed by \cite{liu2016ssd}, which detects multi-scale objects using anchor boxes of different scales and aspect ratios applied across the units of the feature map.Throughout the years, the YOLO algorithm has evolved through multiple versions including YOLOv2 \cite{redmon2017yolo9000}, YOLOv3 \cite{redmon2018yolov3}, YOLOv4 \cite{bochkovskiy2020yolov4}, YOLOv6 \cite{li2022yolov6}, YOLOv7 \cite{wang2023yolov7}, YOLOv9 \cite{wang2024yolov9}, and YOLOv10 \cite{wang2024yolov10}, each aiming to enhance both detection accuracy and speed.In recent years, the YOLO algorithm, with its efficient and real-time object detection capabilities, has been widely applied across multiple fields. Its applications span areas such as autonomous driving\cite{li2022cross}\cite{sindhwani2021comparative}, video surveillance\cite{nguyen2021yolo}\cite{xu2021surveillance}, medical image analysis\cite{ragab2024comprehensive}, robotic vision\cite{cao2021detecting}, and smart agriculture\cite{wu2020using}\cite{al2022yolo}. The rapid development and iteration of the YOLO algorithm have driven significant advancements in object detection technology, not only improving detection accuracy and speed but also enhancing robustness in complex scenarios. This has made YOLO one of the most influential and popular algorithms in the field of computer vision, greatly promoting the development and technological innovation of related applications.
\begin{figure*}[htbp]
	\centering
	\includegraphics[width=\textwidth]{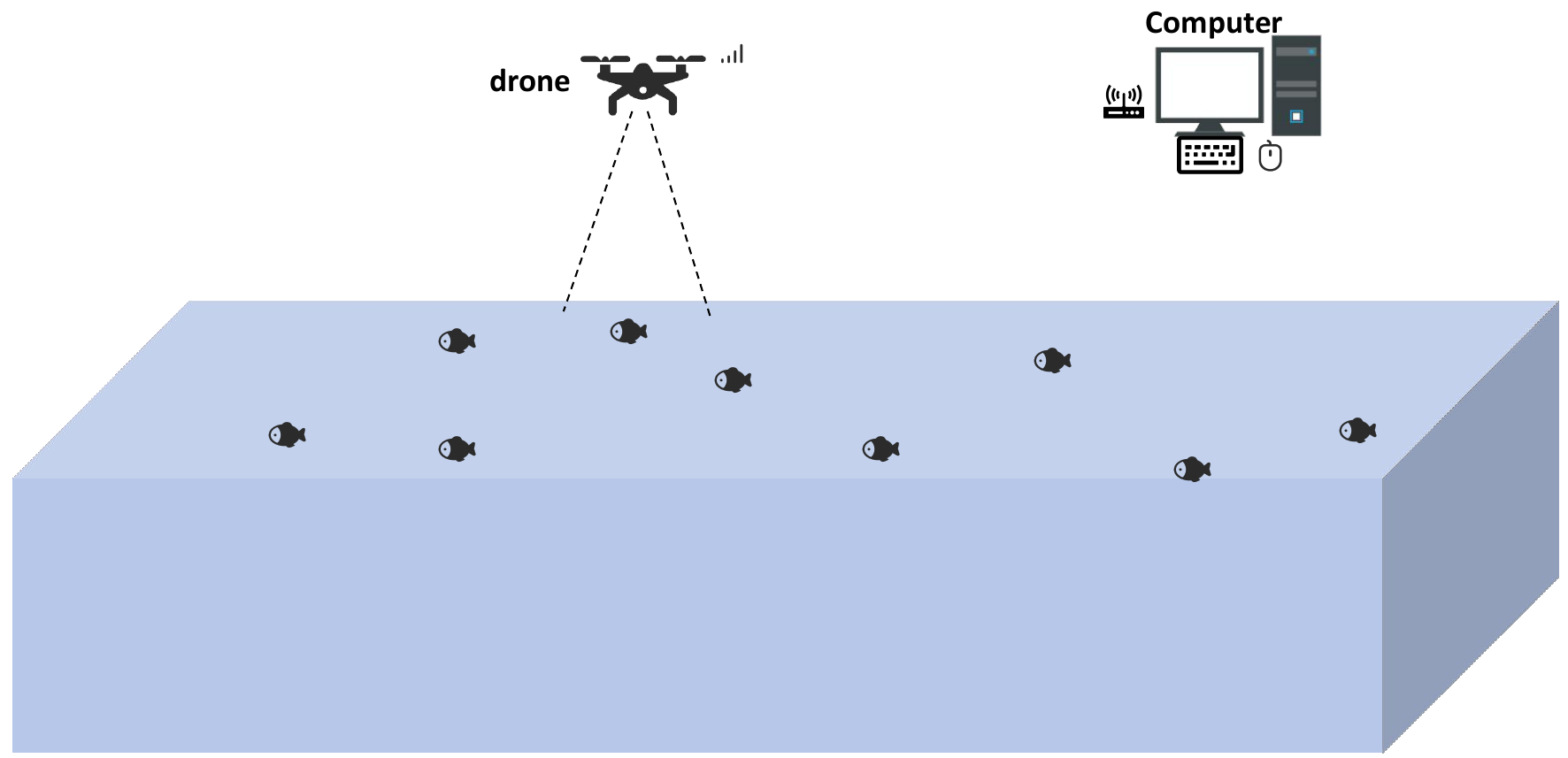}
	\caption{Experimental data collection system.}
	\label{fig:drone}
\end{figure*}
\begin{figure*}[htbp]
	\centering
	\includegraphics[width=\textwidth]{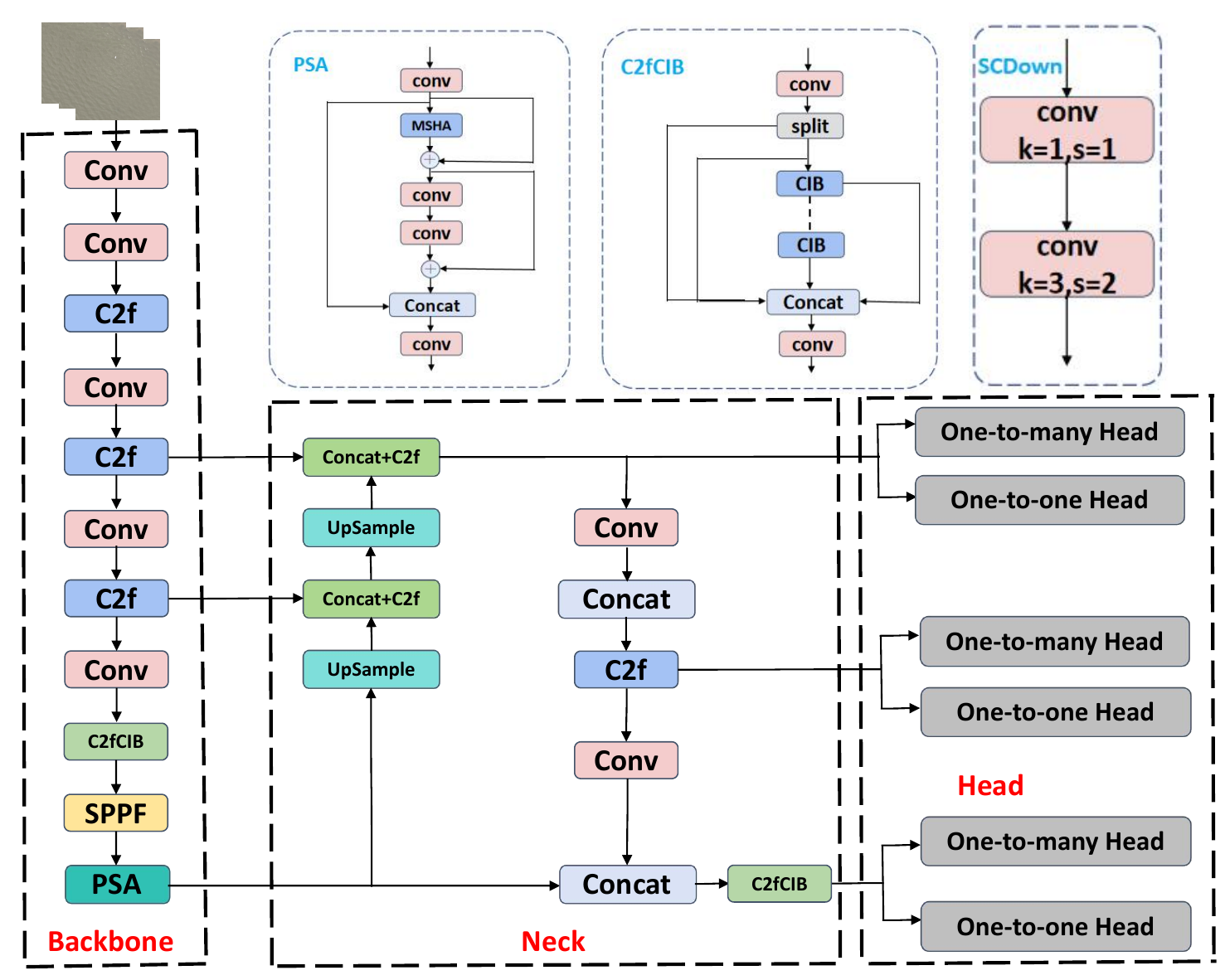}
	\caption{Structure diagram of YOLOv10.}
	\label{fig:yolov10}
\end{figure*}
\begin{figure*}[!h]
	\centering
	\subfloat[Convolution]{\includegraphics[width=2.2in]{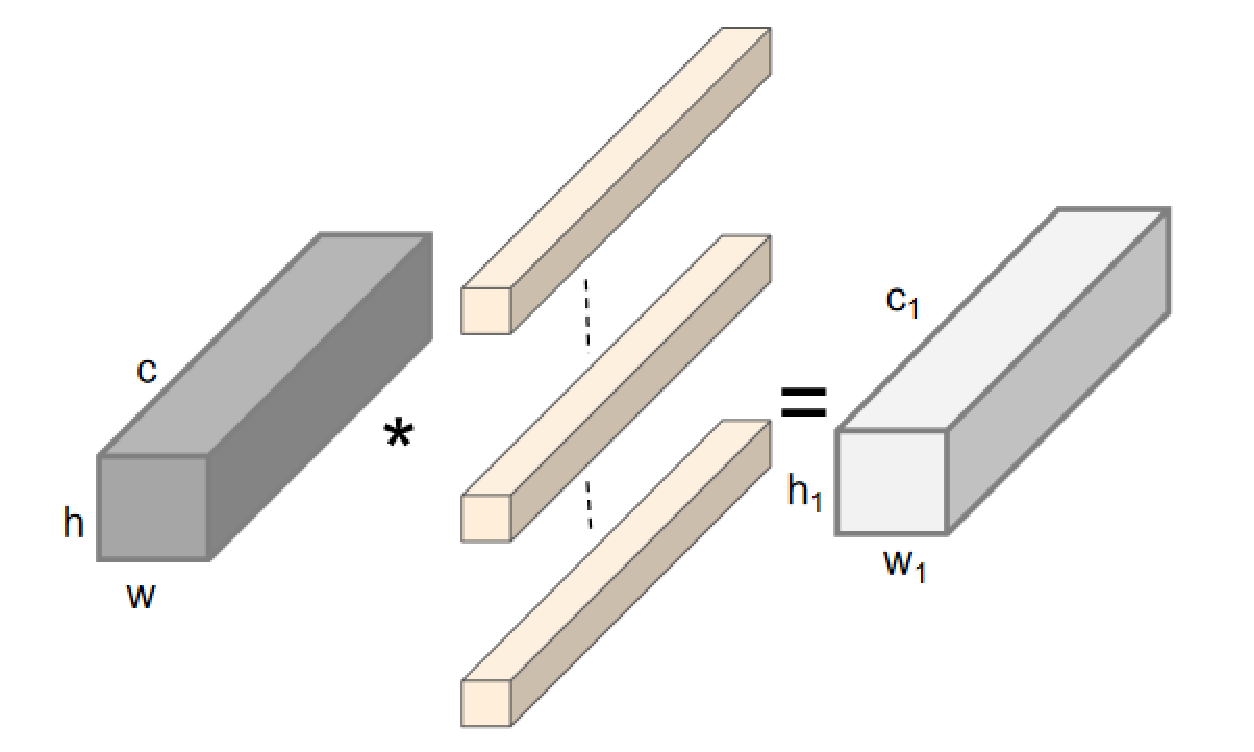} \label{convX}}
	\hfill
	\subfloat[Depthwise/Group Convolution]{\includegraphics[width=2.2in]{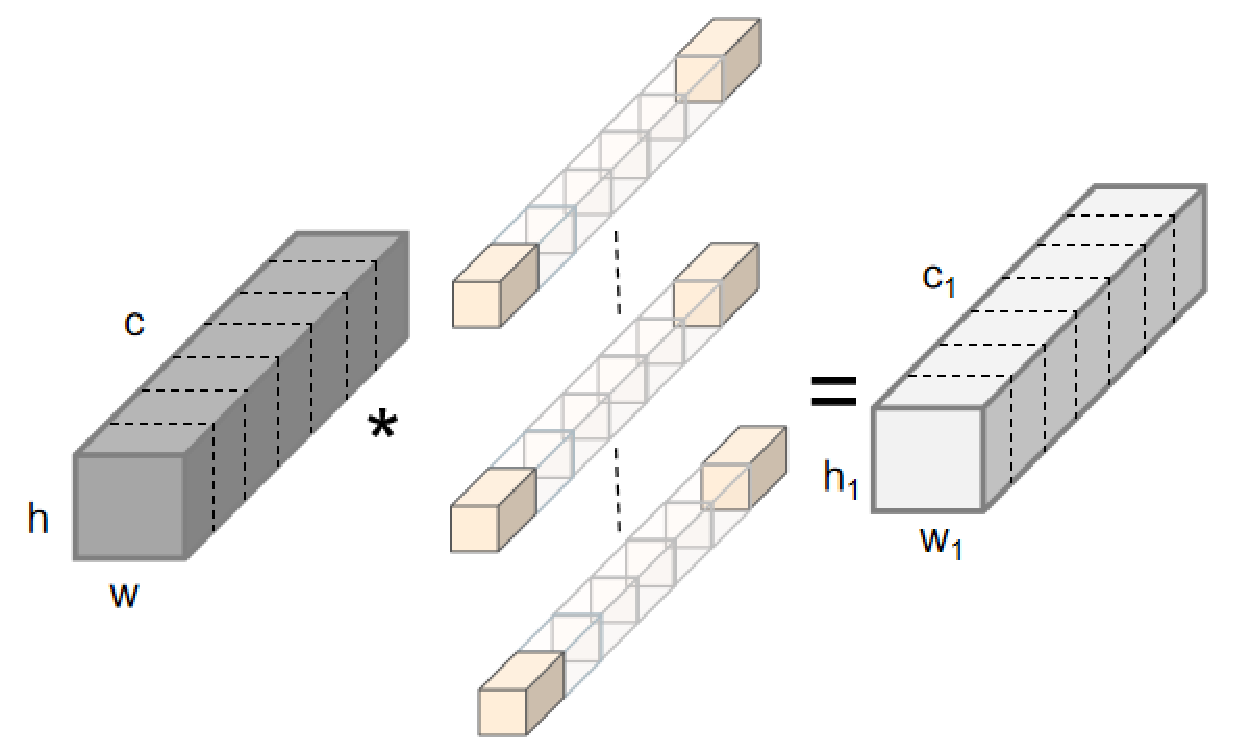} \label{convY}}
	\hfill
	\subfloat[Partial Convolution]{\includegraphics[width=2.2in]{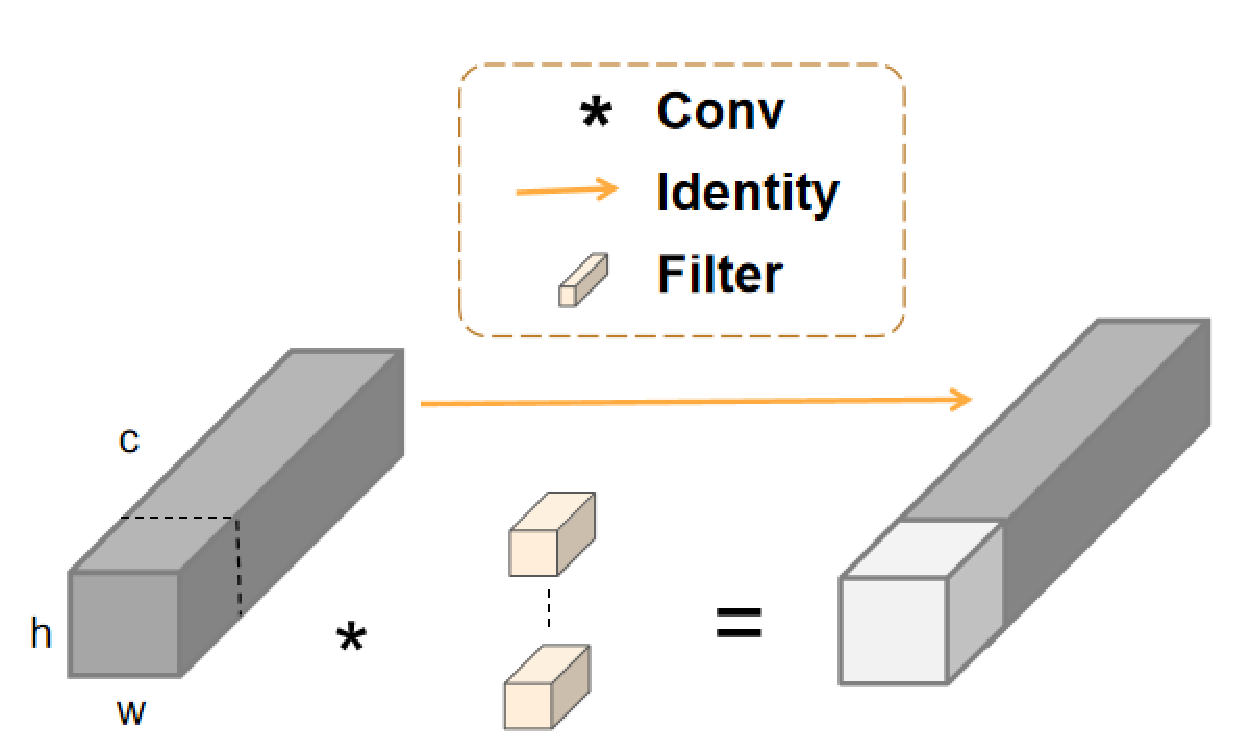} \label{convZ}}
	\caption{Illustrations of different convolution operations. (a) Standard Convolution: A filter is applied across the entire input feature map to produce an output feature map. (b) Depthwise/Group Convolution: Depthwise convolution applies a single filter per input channel, and group convolution divides input channels into groups, applying separate filters within each group. (c) Partial Convolution: The convolution operation is applied only to the unmasked regions of the input, effectively reducing memory access times.}
	\label{fig:conv}
\end{figure*}
\begin{figure}[h!]
	\centering
	\includegraphics[width=1\linewidth]{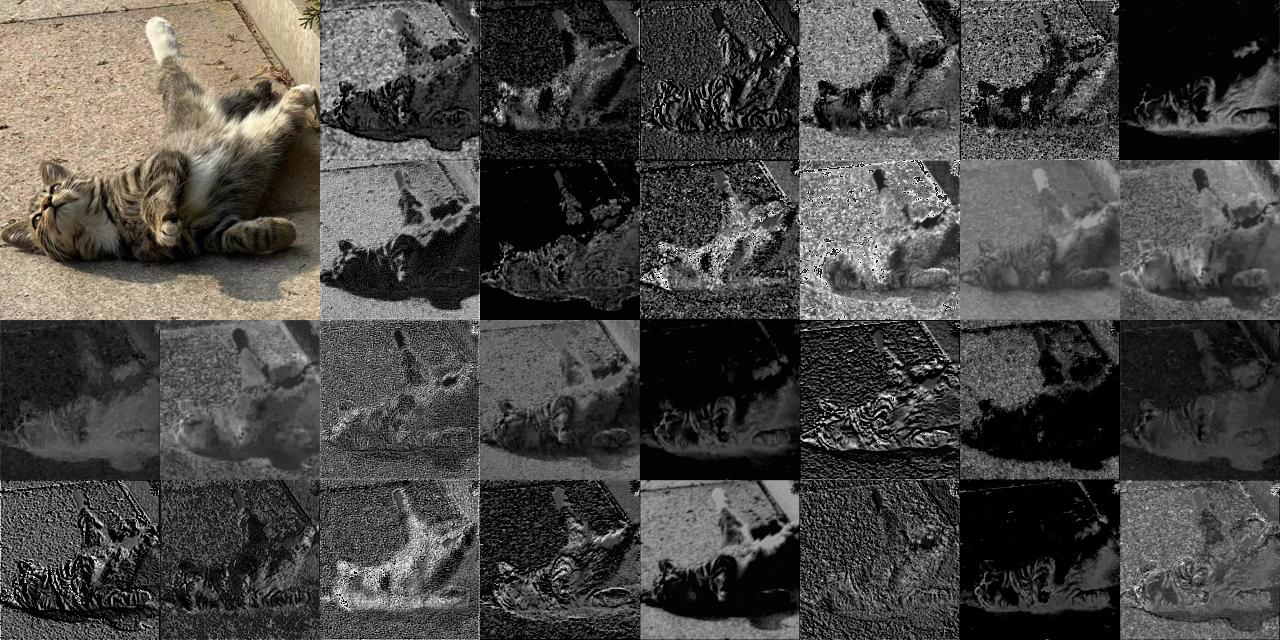}
	\caption{Visualization of feature maps in an intermediate layer of a pre-trained ResNet50, with the top-left image as the input. Qualitatively, we can see the high redundancies across different channels.}
	\label{fig:cat}
\end{figure}
\section{Materials and methods}
\subsection{Data set acquisition}
The research data were collected from the aquaculture lake located at Junshan Fishery Group Co., Ltd., in Yueyang, Hunan. During data collection, a drone was utilized to capture videos of dead fish from different heights and angles over the lake surface (as depicted in \hyperref[fig:drone]{Figure 1}). The videos were recorded at a resolution of 3840 × 2160 pixels, capturing at a rate of 60 frames per second. Subsequently, the videos were segmented into individual frames, and appropriate images were selected to compile the subsequent target detection dataset.
\begin{figure*}[htbp]
	\centering
	\includegraphics[width=\textwidth]{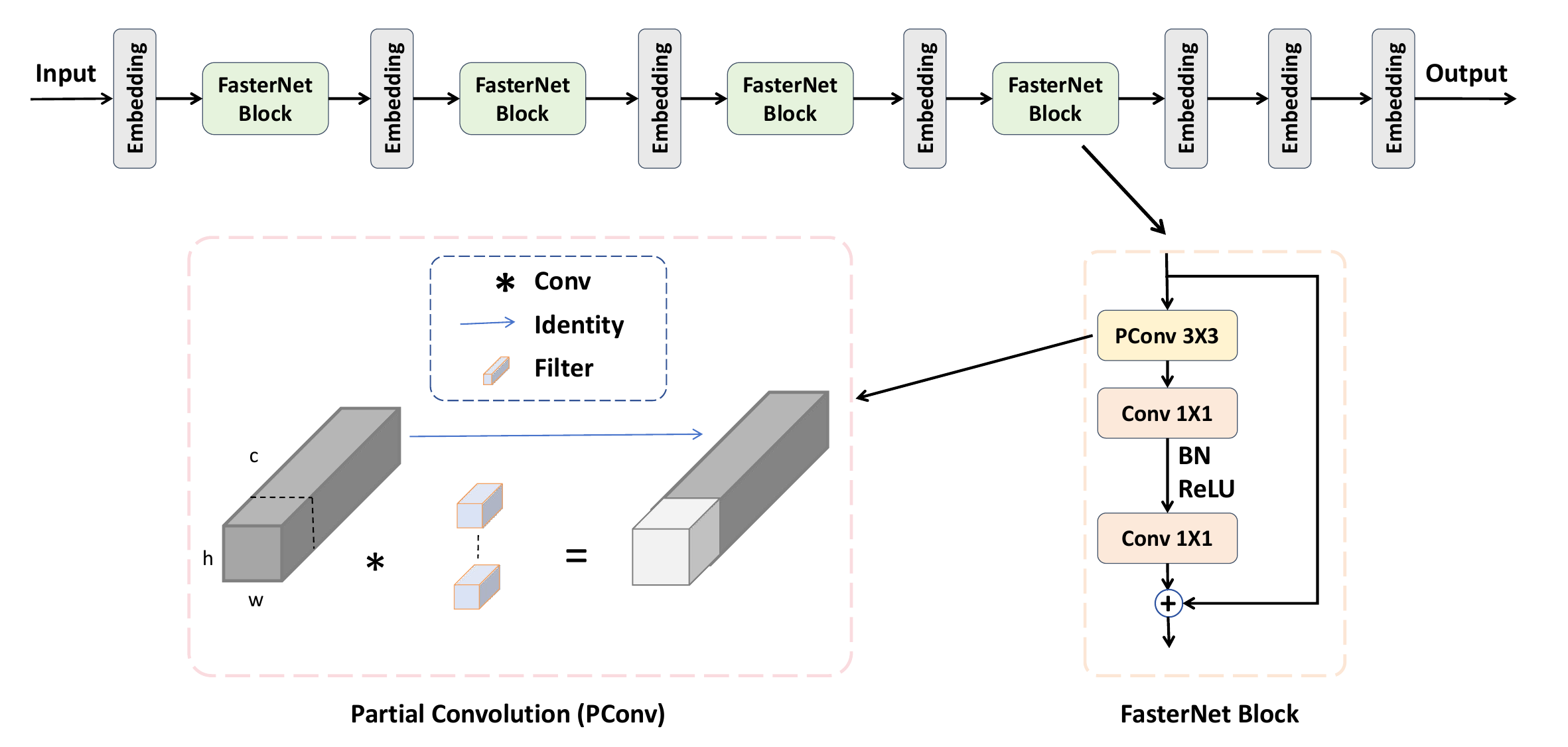}
	\caption{Structure of FasterNet.}
	\label{fig:fasterNet}
\end{figure*}
\begin{figure*}[tbph]
	\centering
	\begin{tikzpicture}
		\node[anchor=south west,inner sep=0] (image) at (0,0) {\includegraphics[width=\textwidth]{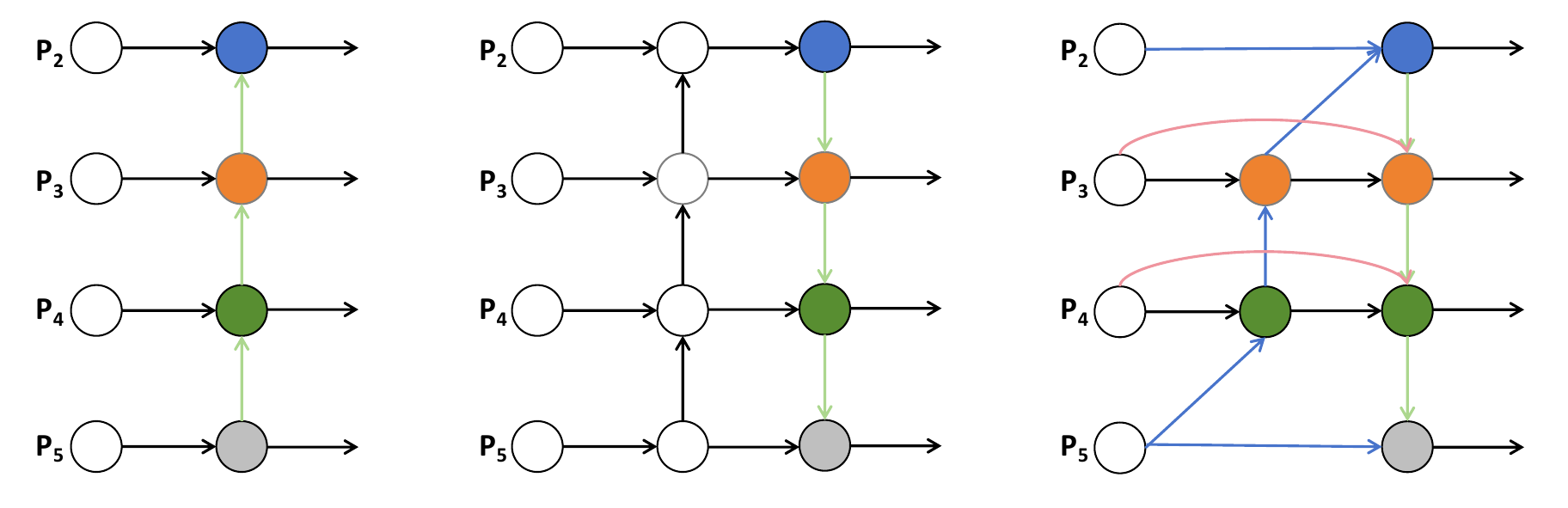}};
		\begin{scope}[x={(image.south east)},y={(image.north west)}]
			\node at (0.13, 0.0005) {(a) FPN};
			\node at (0.45, 0.0005) {(b) PANet};
			\node at (0.83, 0.0005) {(c) BiFPN};
		\end{scope}
	\end{tikzpicture}
	\caption{Feature network design.(a) FPN introduces a top-down pathway to integrate multi-scale features from different layers; (b) PANet builds upon FPN by adding a bottom-up pathway; (c) BiFPN removes nodes with only one input edge and introduces an additional pathway between input and output nodes at the same layer.}
	\label{fig:FPN}
\end{figure*}
Utilizing high-resolution video capture through drones enabled thorough coverage of the lake surface, facilitating the acquisition of a comprehensive dataset of dead fish images. This approach significantly enhances the accuracy and reliability of the detection model. Furthermore, capturing images from various heights and angles contributed to a diverse dataset, enhancing the model's adaptability to complex environmental conditions. Following careful selection and processing, these images provided ample material for both training and testing the model, ensuring the scientific rigor and effectiveness of the experiment.

A total of 500 images containing dead fish were initially collected. Utilizing data augmentation techniques such as image flipping, scaling, segmentation, and translation,the dataset size was notably expanded to 1050 images. The images were allocated into three sets: 600 for training, 200 for validation, and 250 for testing.Employing data augmentation not only increased the dataset size but also enhanced the model's capacity to generalize, ensuring consistently high detection accuracy across various environments and conditions. This augmented dataset enriched the training samples, thereby enhancing the reliability and stability of detection outcomes.

In this experiment, we utilized the image annotation tool X-AnyLabeling (\href{https://github.com/CVHub520/X-AnyLabeling}{github.com/CVHub520/X-AnyLabeling}) to accurately annotate the dataset images. Annotation produced txt files containing target types and coordinate information crucial for training the target detection algorithm. This ensured accurate identification and localization of dead fish targets in each image.

\subsection{The proposed FN-YOLO}

In previous versions of YOLO, a one-to-many label assignment strategy was commonly used during training, where one ground truth object corresponded to several positive samples.While this approach can enhance performance, it necessitates the use of Non-Maximum Suppression (NMS) \cite{neubeck2006efficient} during inference to select the best positive predictions. This requirement not only decreases inference speed but also causes performance to be dependent on the hyperparameters of NMS, thus complicating the end-to-end deployment of YOLO.

NMS is a widely adopted post-processing technique in object detection algorithms. It serves to minimize redundant bounding boxes and uphold the accuracy of detection outcomes. The main objective of NMS is to retain the best detection box for the same object while suppressing lower-scoring overlapping boxes. The working principle of NMS is as follows:
\begin{enumerate}
	\item \textbf{Sorting Detection Boxes: }First, sort all detection boxes by confidence score (i.e., detection score) in descending order.
	\item \textbf{Choosing the detection box with the highest score:}Choose the detection box with the highest score as the current best detection result.
	\item \textbf{Calculating Overlapping Areas: }For the remaining detection boxes, calculate their overlap with the current best detection box, typically using the Intersection over Union (IoU) metric.
	\item \textbf{Suppressing Overlapping Boxes: }Suppress (i.e., remove) those detection boxes whose overlap with the current best detection box exceeds a certain threshold, as they are likely duplicate detections of the same object.
	\item \textbf{Repeating Steps 2-4: }Repeat the above steps for the remaining detection boxes until no boxes are left.
\end{enumerate}

In YOLOv10\cite{wang2024yolov10}, the authors proposed a no NMS training strategy, which achieves high efficiency and competitive performance through dual-label assignment and consistent matching metrics. This strategy combines the advantages of one-to-many label assignment (assigning multiple predicted bounding boxes to each true bounding box as positive samples) and one-to-one label assignment (assigning only one predicted bounding box to each true bounding box as a positive sample), using these two methods respectively during training and inference. Specifically, the authors added another one-to-one head to YOLO, which retained the same structure as the original one-to-many branch and utilized the same optimization objective, but obtained label assignment through one-to-one matching. During the training phase, the model and the two heads undergo joint optimization. This allows the backbone and neck to gain from the extensive supervision afforded by the one-to-many assignment. In the inference phase, the one-to-many head is removed, and predictions are generated using the one-to-one head. This approach permits end-to-end deployment of YOLO without incurring extra inference costs.

In the process of label assignment, both the one-to-one and one-to-many methods employ a metric to quantitatively evaluate the degree of alignment between predictions and instances. To facilitate prediction-aware matching for both branches, the authors developed a unified matching metric, specifically:
\begin{equation}
m(\alpha, \beta) = s \cdot p^{\alpha} \cdot \text{IoU}(\hat{b}, b)^{\beta}
\label{eq:eq1}
\end{equation}
In the formula presented by \hyperref[eq:eq1]{Equation (1)}, \( p \) denotes the classification score, while \( \alpha \) and \( \beta \) represent the bounding boxes for the prediction and the instance, respectively. The spatial prior, \( s \), indicates whether the prediction's anchor point is located within the instance. The hyperparameters \( \alpha \) and \( \beta \) are crucial for balancing the semantic prediction task and the position regression task. We define the metrics for one-to-many and one-to-one assignments as \( m_{o2m} = m(\alpha_{o2m}, \beta_{o2m}) \) and \( m_{o2o} = m(\alpha_{o2o}, \beta_{o2o}) \), respectively. The author illustrates that the supervision gap between the two branches can be measured by calculating the 1-Wasserstein distance of the different classification objectives, as shown in \hyperref[eq:eq2]{Equation (2)}.
\begin{equation}
	A = t_{o2o,i} - I(i \in \Omega) \cdot t_{o2m,i} + \sum_{k \in \Omega\backslash\{i\}} t_{o2m,k}
	\label{eq:eq2}
\end{equation}

As \(t_{o2m,i}\) increases, the supervision gap gradually decreases, reaching its minimum when \(t_{o2m,i} = u^*\), indicating that \(i\) is the best positive sample in \(\Omega\). To achieve this, the author proposes a consistent matching metric, namely \(\alpha_{o2o}=r\cdot\alpha_{o2m}\) and \(\beta_{o2o}=r\cdot\beta_{o2m}\), implying \(mo2o=mr o2m\). Certainly! Here's the revised version of the text:

Thus, the optimal positive sample for the one-to-many head is equally suitable for the one-to-one head. As a result, both heads can be optimized in a consistent and harmonious manner. For the sake of simplicity, we assume \(r=1\), i.e., \(\alpha_{o2o}=\alpha_{o2m}\) and \(\beta_{o2o}=\beta_{o2m}\).

The YOLOv10 algorithm is developed based on YOLOv8, and its main network structure includes the backbone, neck, and head. The architecture of YOLOv10 is illustrated in \hyperref[fig:yolov10]{Figures 2}. The backbone network consists of multiple layers, including convolutional layers, C2f modules, SCDown modules, SPPF modules, and PSA modules. These layers are responsible for extracting features and downsampling operations from input images to generate feature maps of different resolutions and semantic levels. The neck part adopts the traditional PAN (Path Aggregation Network)\cite{liu2018path} structure, which includes a bottom-up feature pyramid to transmit strong localization features, and a feature pyramid from the top down to deliver robust semantic features.

The YOLOv10 network model comprises six versions: v10N, v10S, v10M, v10B, v10L, and v10X, differing mainly in width and depth. Although the v10N model boasts the highest detection speed, its detection accuracy tends to be relatively lower when dealing with small targets or objects affected by background interference. This phenomenon stems from the original model's lack of a specialized detection layer for tiny targets and its relatively weak capability to extract and fuse effective information features. To tackle this issue, a specialized layer for detecting small targets can be designed,along with optimizations to the feature fusion method in the neck network to enhance interaction and fusion among features. However, adding a small target detection layer always results in a notable increase in the model's parameter count. To mitigate this, we implement a strategy by substituting the original backbone network with Fasternet, which reduces the model's parameter count while achieving a trade-off between detection speed and accuracy.

\begin{figure*}[tbph]
	\centering
	\begin{tikzpicture}
		\node[anchor=south west,inner sep=0] (image) at (0,0) {\includegraphics[width=\textwidth]{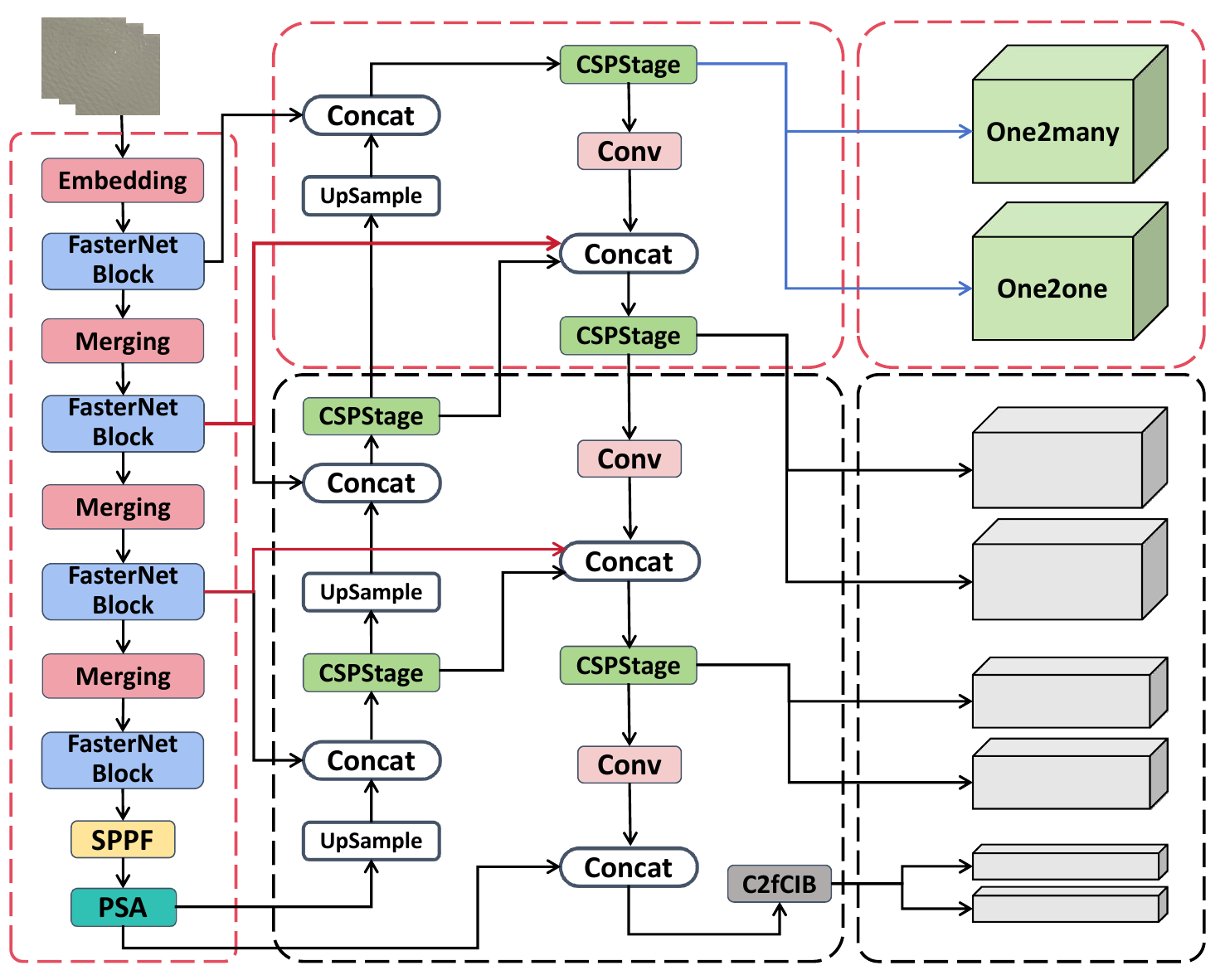}};
		\begin{scope}[x={(image.south east)},y={(image.north west)}]
			\node at (0.10, 0.0005) {(a) Backbone};
			\node at (0.45, 0.0005) {(b) Neck};
			\node at (0.83, 0.0005) {(c) Head};
		\end{scope}
	\end{tikzpicture}
	\caption{Structure diagram of FN-YOLO. (a) Backbone: Replaced with FasterNet while retaining YOLOv10's SPPF and PSA modules for better feature extraction. (b) Neck: Modified feature fusion by adding a new connection between input and output nodes at the same layer, replacing the original C2f module with CSPStage, and introducing additional upsampling and downsampling modules to better utilize and fuse feature information. (c) Head: Added a new small object detection head to enhance the detection of small objects.}
	\label{fig:FN-yolo}
\end{figure*}

Therefore, this paper introduces FN-YOLO, an enhanced target detection model tailored for the specific task of identifying deceased fish floating on the water's surface. The overall structure of FN-YOLO is illustrated in \hyperref[fig:FN-yolo]{Figure 7}. The main improvements include: (1) replacing the backbone of the original model with Fasternet, which effectively decreases the number of parameters in the model and enhances computational efficiency. (2) Improvements in feature fusion, including the addition of cross-layer connections and the replacement of the original C2f module with the CSPStage module. These improvements notably boost the model's capability to leverage features across various scales, further strengthening the efficiency of information transmission and utilization in features. Therefore, the model can better capture information from different levels and achieve more comprehensive information interaction during feature fusion, thereby enhancing its ability to recognize and detect targets. (3) Incorporating a specialized layer for detecting small targets enhances the model's accuracy in identifying and locating small objects, leading to improved overall performance.

\subsubsection{Fasternet}
The FasterNet \cite{chen2023run} enables model developers to choose a visual model that aligns with the resource limitations of their applications.By incorporating Partial Convolution (PConv) and Pointwise Convolution (PWConv), the number of operations and memory usage is significantly reduced, thereby achieving an effective balance between latency and model accuracy.The main advantages of FasterNet include high computational efficiency, low latency, and hierarchical design, making it suitable for various scenarios requiring fast processing and efficient computation.

(a) Depthwise Convolution. 

The standard convolution operation is a commonly used technique in image processing and computer vision. As shown in \hyperref[convX]{Figure 3(a)}, it uses a small filter (convolution kernel) that slides over the input feature map, performing element-wise multiplication and summation for each local region to generate an output feature map. This operation effectively extracts features from the input data through local perception, parameter sharing, and translation invariance. Depthwise Convolution, a variation of the convolution operation, is widely used in various neural networks. For an input tensor \( I \in \mathbb{R}^{c \times h \times w} \), DWConv employs \( c \) convolutional kernels \( W \in \mathbb{R}^{k \times k} \) independently across each input channel, producing an output tensor \( O \in \mathbb{R}^{c \times h \times w} \). As illustrated in \hyperref[convY]{Figure 3(b)}, each filter slides over a single input channel to generate a corresponding output channel.This method greatly decreases the quantity of floating-point operations (FLOPs), decreasing from \( h \times w \times k^2 \times c^2 \) required by standard convolution to \( h \times w \times k^2 \times c \).

Although DWConv is highly efficient in reducing FLOPs, it is typically followed by a Pointwise Convolution (PWConv) to compensate for potential accuracy loss. Directly replacing a standard convolution with DWConv can lead to noticeable performance degradation. To address this, the number of channels \( c' \) (where \( c' > c \)) is often increased, as seen in the Inverted Residual Block, where the number of channels can be increased up to six times. However, this increase also raises memory access requirements, introducing latency that can slow down the overall computation process, especially on I/O-bound devices.The frequency of memory accesses escalates to
\begin{equation}
	h \times w \times 2c' + k^2 \times c' \approx h \times w \times 2c'
\end{equation}
while standard convolution requires 
\begin{equation}
	h \times w \times 2c + k^2 \times c^2 \approx h \times w \times 2c
\end{equation}
Additionally, the memory accesses here are primarily for I/O operations, making further optimization challenging.

(b) Partial convolution. 

In light of the aforementioned limitations of DWConv, Partial Convolution (PConv) has emerged.As shown in \hyperref[convZ]{Figure 3(c)},the core idea of PConv is to apply standard convolution operations to only a subset of channels in the input feature map to extract feature information, while keeping the remaining channels unchanged. This method is grounded in the observation that various channels of the feature map often show a high degree of similarity, as shown in \hyperref[fig:cat]{Figures 4}. When the input and output feature maps possess an identical number of channels, the computational cost (FLOPs) of PConv is calculated as:
\begin{equation}
h \times w \times k^2 \times c^2_p.
\end{equation}
When \( r = \frac{c_p}{c} = \frac{1}{4} \), the FLOPs of PConv decrease to \( \frac{1}{16} \) of a regular convolution. PConv also requires less memory access, i.e.,
\begin{equation}
	h \times w \times 2c_p + k^2 \times c^2_p \approx h \times w \times 2c_p.
\end{equation}
Where \( c \) represents the number of channels in the feature map, \( c_p \) denotes the number of channels in the aggressive convolution operation, and \( k \) represents the size of the convolutional kernel.

(c) The proposal of FasterNet. 

Based on PConv and PWConv, the researchers proposed FasterNet, a novel neural network. It boasts exceptional running speed and is highly effective for visual tasks. The overall structure, as shown in \hyperref[fig:fasterNet]{Figure 5}, comprises four stages. Each stage is preceded by a standard convolution layer with a stride of 4 or 2, used for downsampling or increasing the number of channels. Each stage contains multiple FasterNet blocks, where each block is composed of one PConv layer succeeded by two PWConv layers.To preserve feature diversity while accelerating inference, the authors added normalization layers and activation functions between every two intermediate PWConv layers. Additionally, the network employs BN for normalization, as it can fuse with adjacent convolution layers, further reducing latency.
\begin{figure*}[tbph]
	\centering
	\begin{tikzpicture}
		\node[anchor=south west,inner sep=0] (image) at (0,0) {\includegraphics[width=\textwidth]{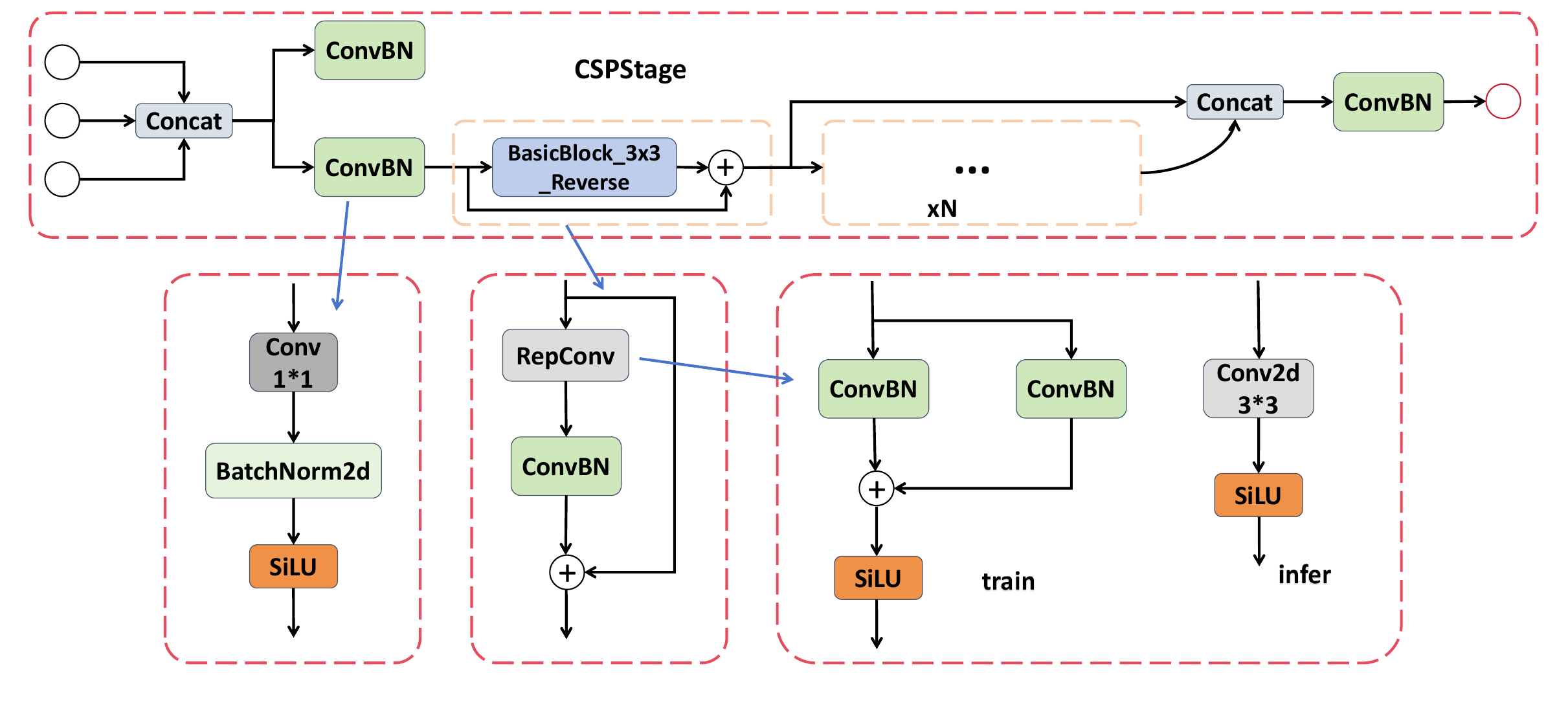}};
		\begin{scope}[x={(image.south east)},y={(image.north west)}]
			\node at (0.18, 0.0005) {(a) ConvBN};
			\node at (0.38, 0.0005) {(b) BasicBlock\_3x3\_Reverse};
			\node at (0.68, 0.0005) {(c) RepConv};
		\end{scope}
	\end{tikzpicture}
	\caption{Structure diagram of CSPStage.}
	\label{fig:CSPStage}
\end{figure*}
\begin{figure*}[htbp]
	\centering
	\includegraphics[width=\textwidth]{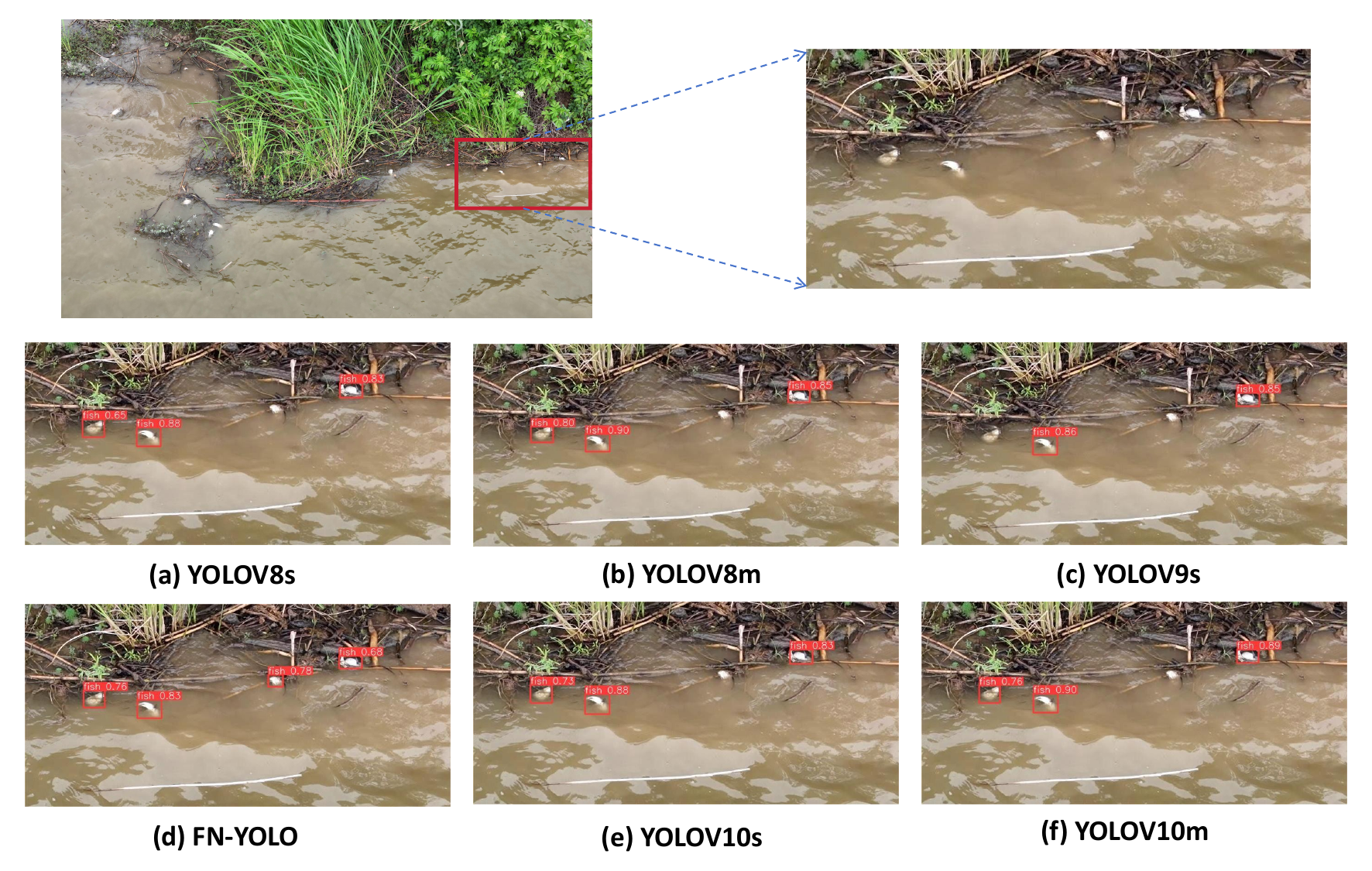}
	\caption{The comparison diagram of the results of different target detection models.}
	\label{fig:reaultvisual}
\end{figure*}
\subsubsection{The structure and design of Neck}

(a) Feature Pyramid Networks. 

FPN (Feature Pyramid Network) \cite{lin2017feature} is an architecture designed to enhance the feature extraction capabilities of deep learning models. As depicted in \hyperref[fig:FPN]{Figure 6(a)}, FPN incorporates a top-down pathway and lateral connections to combine feature maps across various levels, forming a multi-scale feature pyramid. This significantly improves the model's performance in object detection and image segmentation tasks, particularly for detecting and segmenting objects of various scales.

PANet (Path Aggregation Network) \cite{liu2018path}, depicted in \hyperref[fig:FPN]{Figure 6(a)}, enhances the effectiveness of object detection and instance segmentation. Building upon FPN, PANet introduces a bottom-up path aggregation mechanism that strengthens feature map fusion, enhancing feature richness and robustness. This capability enhances the model's ability to capture information across multiple scales. PANet's bidirectional feature fusion and other innovative modules notably enhance detection accuracy and segmentation quality, particularly in handling multi-scale and complex scenes.

BiFPN (Bidirectional Feature Pyramid Network) \cite{tan2020efficientdet}, illustrated in \hyperref[fig:FPN]{Figure 6(c)}, is an optimized feature pyramid network that improves the flow and fusion of information between feature maps at different levels through bidirectional feature fusion and weighted feature fusion mechanisms. It enhances model performance in multi-scale object detection and instance segmentation tasks, offering efficient feature fusion, bidirectional information flow, and flexibility, thus demonstrating wide applicability in high-performance computer vision tasks.
BiFPN is an improved feature pyramid network structure. BiFPN differs from PANet by removing single-input nodes and introducing new connections between input and output nodes within the same layer.It also introduces a learnable weighting mechanism, where different weights are assigned to each input feature during feature fusion. This allows the network to dynamically adjust the contribution of different scale features, optimizing the feature fusion process. However, weighted fusion requires the introduction of weights at each step of feature fusion, which need to be optimized during training. Although these weights enhance the flexibility of feature fusion, they also increase the computational and storage overhead. This extra computational load can significantly impact the overall efficiency of the model, especially when dealing with large-scale datasets or complex models.
Therefore, we only used BiFPN's node connection method and adopted the simplest Concat method for feature fusion. 

(a) The CSPStage. 

To improve the expressive capability of the fused features, we replaced C2f with the CSPStage\cite{xu2022damo} module. As shown in \hyperref[fig:CSPStage]{Figure 8},the CSPStage module processes input data through two convolutional layers, followed by several \texttt{BasicBlock\_3x3\_Reverse} modules for stepwise feature processing. Additionally, the SPP module can be optionally used to further enhance feature extraction capabilities. Ultimately, all features are concatenated along the channel dimension and then subjected to final feature transformation.

The CSPStage module excels at capturing features at different levels, particularly in handling complex data, allowing for more comprehensive feature extraction and fusion, thereby enhancing the model's representation capability. Through the meticulously designed convolutional blocks and feature fusion mechanisms, this module boosts feature extraction and fusion capabilities while maintaining high computational efficiency. Furthermore, by adjusting the \texttt{split\_ratio} parameter, a better balance between computational overhead and feature extraction effectiveness can be achieved.

In summary, the CSPStage module offers significant advantages in feature extraction and fusion, providing robust support for complex data modeling while optimizing the use of computational resources.The final improved neck structure is illustrated in \hyperref[fig:FN-yolo]{Figure 7(b)}.
\subsubsection{Additional detection layer}
The downsampling process within the backbone network produces feature maps of varying sizes. In the previous section, the improved Neck part, similar to the original YOLOv10 Neck, utilized only three feature maps, resulting in inadequate performance in detecting small targets. Therefore, this section further optimizes this structure by adding an upsampling layer in the Neck part to generate a new feature map. This feature map integrates features from both shallow and deep layers, possessing higher resolution and smaller receptive fields, thereby enabling more precise capture of small target features. In the detection section, four detection layers of varying sizes are introduced to comprehensively detect targets within the image. The final enhanced detection architecture is depicted in \hyperref[fig:FN-yolo]{Figure 7(c)}.
\begin{figure}[htbp]
	\centering
	\includegraphics[width=\columnwidth]{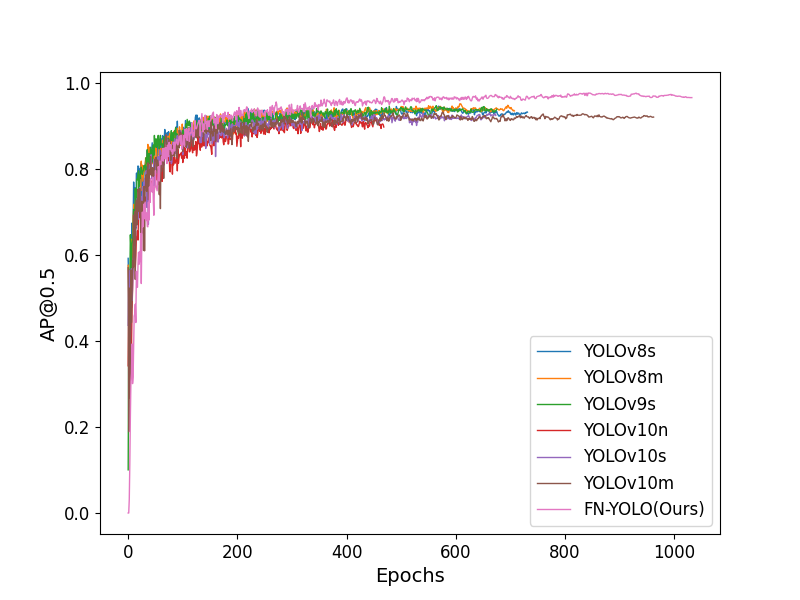}
	\caption{Curves of AP@50 for different detection models.}
	\label{fig:fig10}
\end{figure}
\begin{figure}[htbp]
	\centering
	\includegraphics[width=\columnwidth]{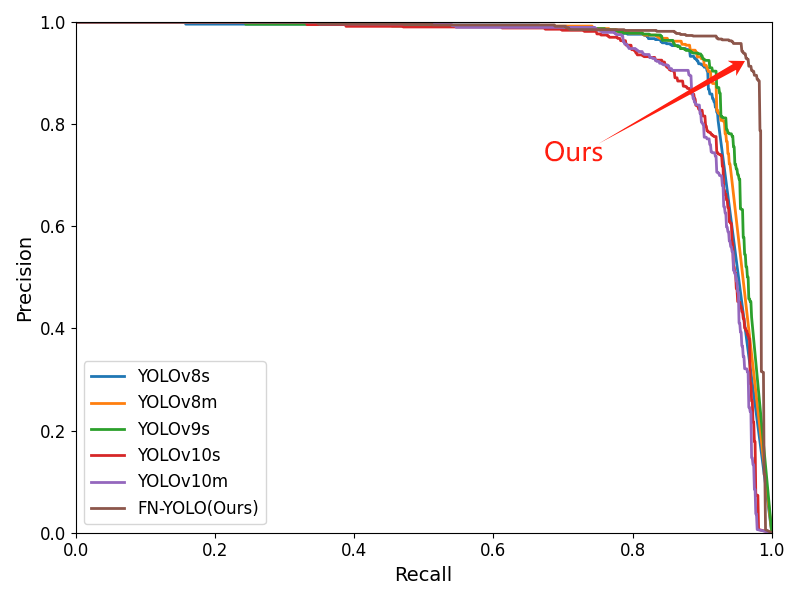}
	\caption{Curves of PR for different detection models.}
	\label{fig:fig11}
\end{figure}
\section{Results}
\subsection{Methods of performance evaluation}
The details of the experimental setup can be found in \hyperref[tab:experimental]{Table 1}. In this experiment, we set the momentum of the learning rate to 0.937 and the initial learning rate to 0.01. The resolution of the input images remains the default 640×640 pixels.Furthermore, the batch size for each training session was configured to 4, totaling 1500 iterations.

The target detection algorithm requires certain indicators to evaluate its performance. In this study, we employ Precision, Recall, and the PR (Precision-Recall) curve, all of which are common metrics used for assessing the performance of target detection algorithms.
\begin{enumerate}
	\item \textbf{Precision:} Precision represents the proportion of targets detected by the algorithm that are truly targets.The precision is computed as TP divided by the sum of TP and FP, represented as
	\begin{equation}
		\text{precision} = \frac{TP}{TP + FP}
	\end{equation}
	\item \textbf{Recall:} Recall indicates the fraction of successfully detected targets out of the total number of targets present. It is computed by dividing TP by the sum of TP and FN, namely,
	\begin{equation}
		\text{recall} = \frac{TP}{TP + FN}
	\end{equation}
\end{enumerate}

\begin{table}[htbp]
	\caption{Experimental configuration.}
	\label{tab:experimental}
	\begin{tabular}{ll}
		\hline
		Configuration  & Parameter \\
		\hline
		CPU & Intel(R) Xeon(R) Gold 6154 \\
		GPU & NVIDIA GeForce RTX 3090 \\
		Operating system & Wondows \\
		Accelerated environment & CUDA 11.1,CUDNN 7.4.1.5 \\
		Development environment & Pycharm2021 \\
		Library & Pytorch 1.7.1 \\
		\hline
	\end{tabular}
\end{table}
Precision and Recall are typically conflicting metrics, where increasing Precision often leads to a decrease in Recall, and vice versa. Therefore, when evaluating the performance of target detection algorithms, it is necessary to balance these two metrics and select the most appropriate threshold or employ other methods to consider them comprehensively. For instance, the PR curve plots the changes in Precision and Recall by altering the decision threshold, aiding in assessing the model's performance at different thresholds. The model performs better when the PR curve approaches the upper right corner.
The area under the PR curve is the Average Precision (AP), which summarizes the model's performance at different thresholds. The calculation formula for \textit{AP} is shown in \hyperref[eq:AP]{Equation (9)}.Specifically, \textit{AP} reflects the average Precision of the model across various Recall levels.The \textit{AP} metric spans from 0 to 1, where superior model performance is indicated by higher values. Typically, AP is used as an evaluation metric for single-class detection. When evaluating the overall performance of multi-class detection, the average of all class \textit{APs} can be computed, known as the mean Average Precision (mAP), as shown in \hyperref[eq:mAP]{Equation (10)}.\textit{AP}$_{50}$ represents the average precision at an IoU (Intersection over Union) threshold of 0.5. IoU quantifies the intersection between the predicted bounding box and the ground truth bounding box.
\begin{equation}
	AP = \int_{0}^{1} Precision \times Recall \, dx \quad
	\label{eq:AP}
\end{equation}

\begin{equation}
	mAP = \frac{\sum_{i=1}^{N} \int_{0}^{1} P(R) \, dR}{N} \quad
	\label{eq:mAP}
\end{equation}

\begin{table*}[h]
	\centering
	\caption{Performance of different algorithms}
	\label{tab:algorithm_performance}
	\begin{tabular}{lcccccc}
		\toprule
		Algorithms & Precision (\%) & Recall (\%) & AP$_{50}$ (\%) & Params (M) & Size (M) & Layers \\
		\midrule
		YOLOv8s & 95.2 & 86.7 & 94.2 & 11.13 & 21.5 & 168 \\
		YOLOv8m & 95.6 & 87.1 & 94.8 & 25.84 & 49.6 & 218 \\
		YOLOv9s & 93.0 & 90.0 & 95.1 & 9.60 & 19.3 & 658 \\
		YOLOv10n & 89.3 & 85.2 & 92.3 & \textbf{2.69} & \textbf{5.7} & 285 \\
		YOLOv10s & 91.6 & 84.6 & 92.6 & 8.04 & 15.8 & 293 \\
		YOLOv10m & 91.4 & 85.2 & 92.7 & 16.45 & 32.0 & 369 \\
		FN-YOLO (Ours) & \textbf{95.7} & \textbf{94.5} & \textbf{97.5} & 2.87 & 6.2 & \textbf{503} \\
		\bottomrule
	\end{tabular}
\end{table*}
\subsection{Results of different detection models}
\begin{figure*}[htbp]
	\centering
	\includegraphics[width=\textwidth]{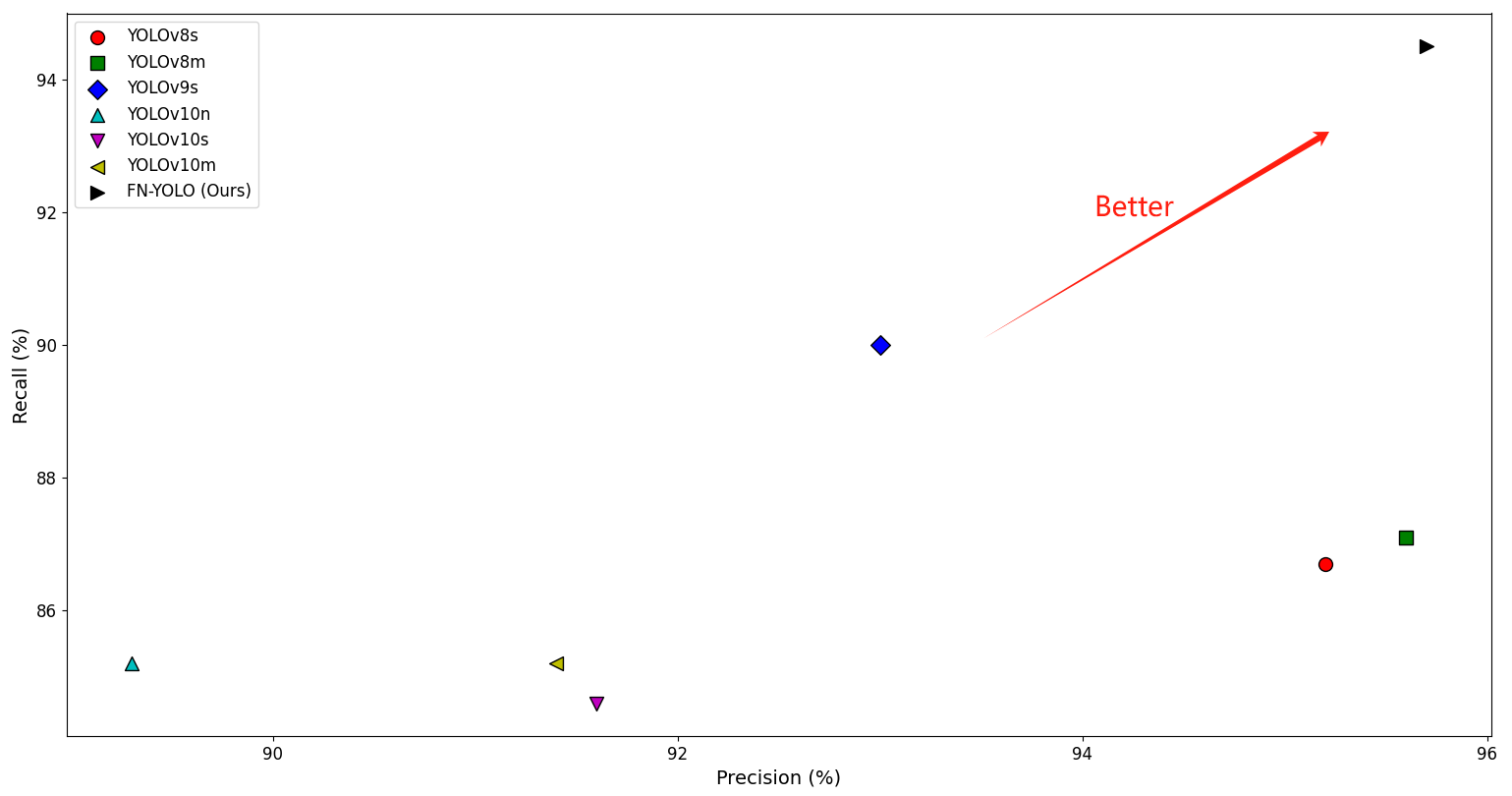}
	\caption{Comparison of Precision and Recall for Different Detection Models.}
	\label{fig:qipao}
\end{figure*}
\begin{figure*}[!h]
	\centering
	\subfloat[Radar Chart of YOLOv8s and FN-YOLO]{\includegraphics[width=3.3in]{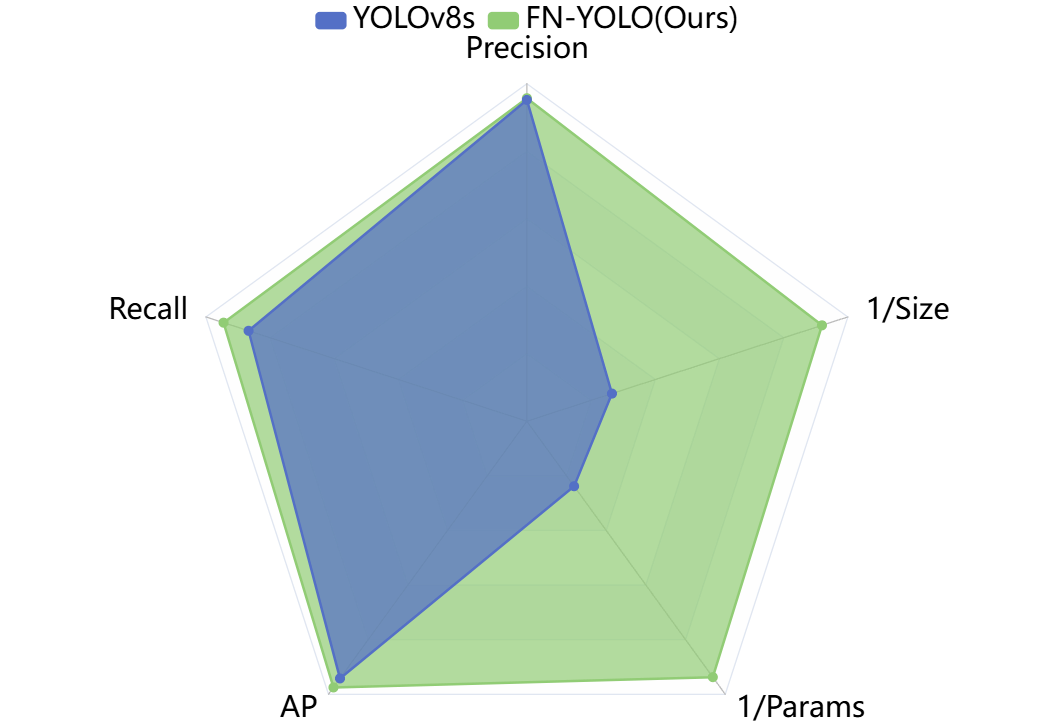} \label{ldX}}
	\hfill
	\subfloat[Radar Chart of YOLOv10s and FN-YOLO]{\includegraphics[width=3.3in]{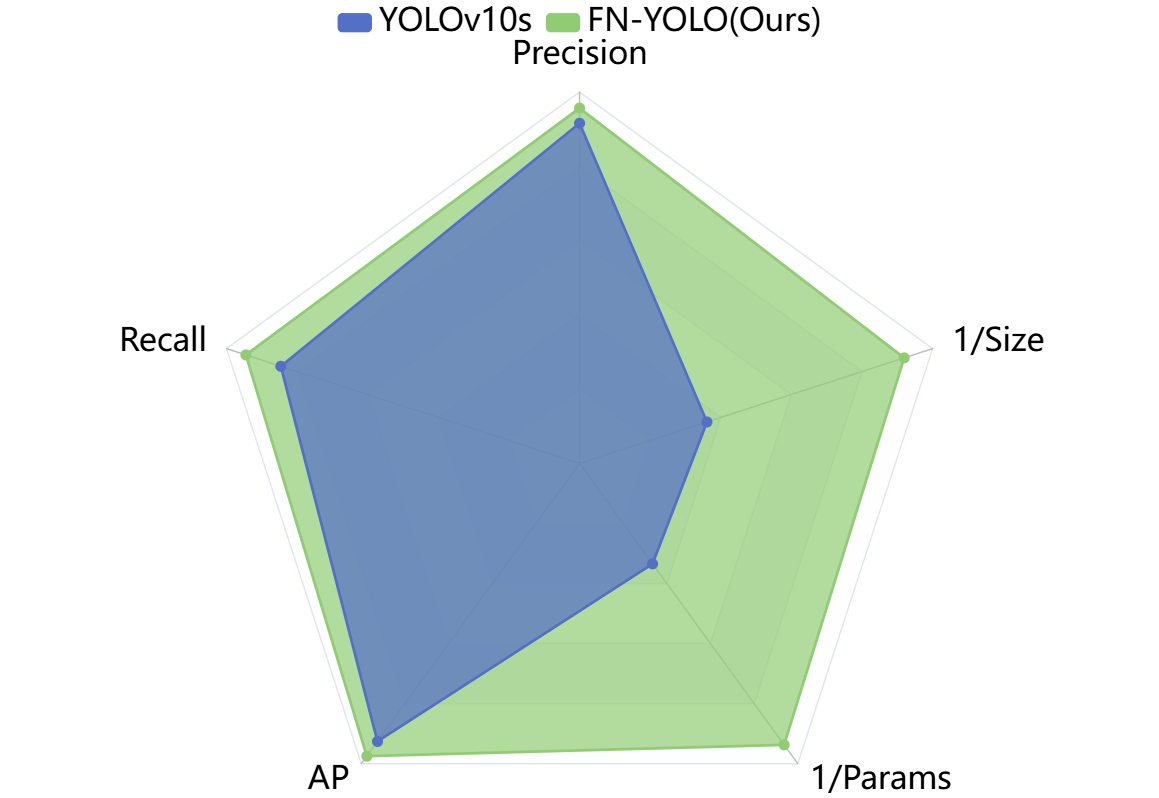} \label{ldY}}
	\hfill
	\caption{Radar charts of YOLOv8s, YOLOv10s, and FN-YOLO. (a) Shows that FN-YOLO outperforms YOLOv8s in all aspects. (b) Shows that FN-YOLO outperforms YOLOv10s in all aspects.}
	\label{fig:ld}
\end{figure*}
\begin{figure*}[!h]
	\centering
	\subfloat[AP@50]{\includegraphics[width=3.3in]{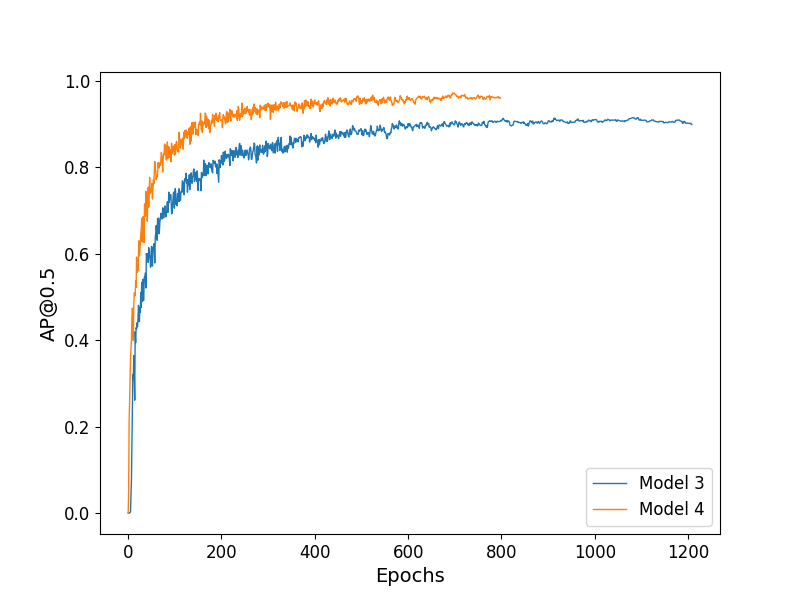} \label{p2w}}
	\hfill
	\subfloat[AP@50-95]{\includegraphics[width=3.3in]{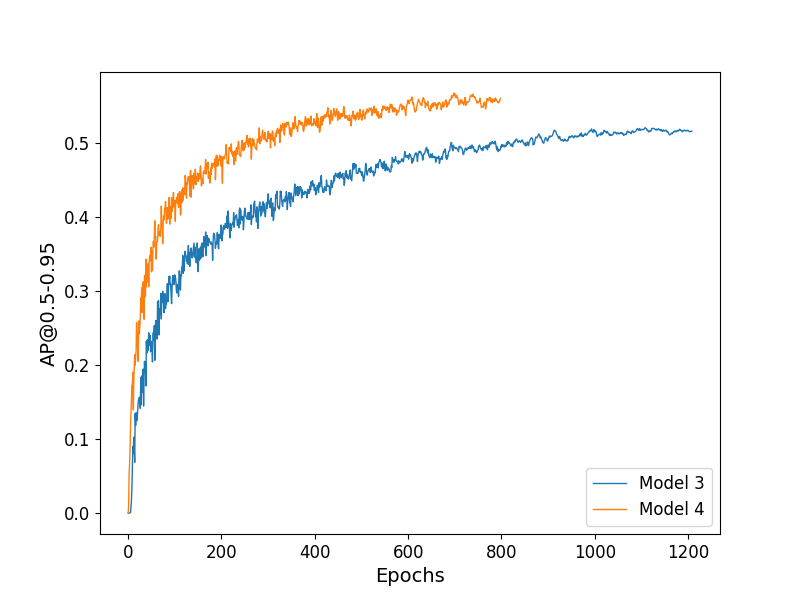} \label{p2x}}
	\vfill
	\subfloat[PR]{\includegraphics[width=3.3in]{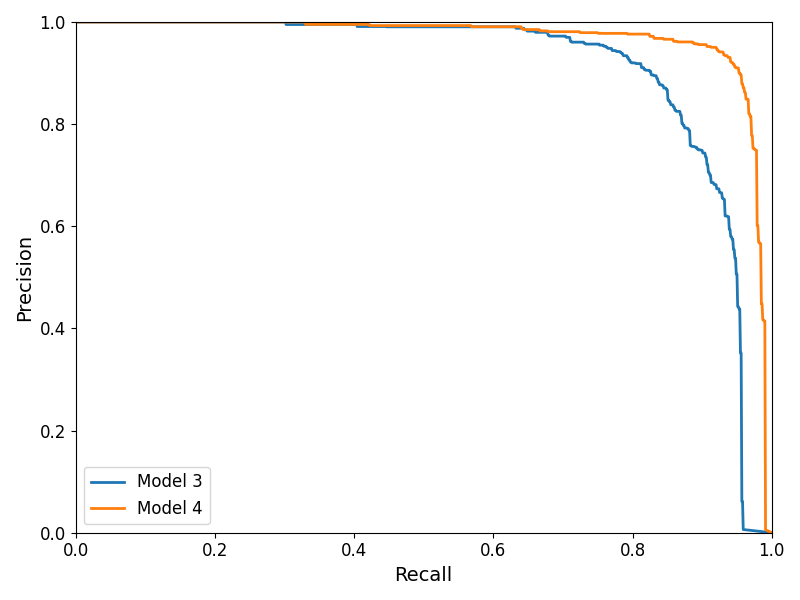} \label{p2y}}
	\hfill
	\subfloat[F1]{\includegraphics[width=3.3in]{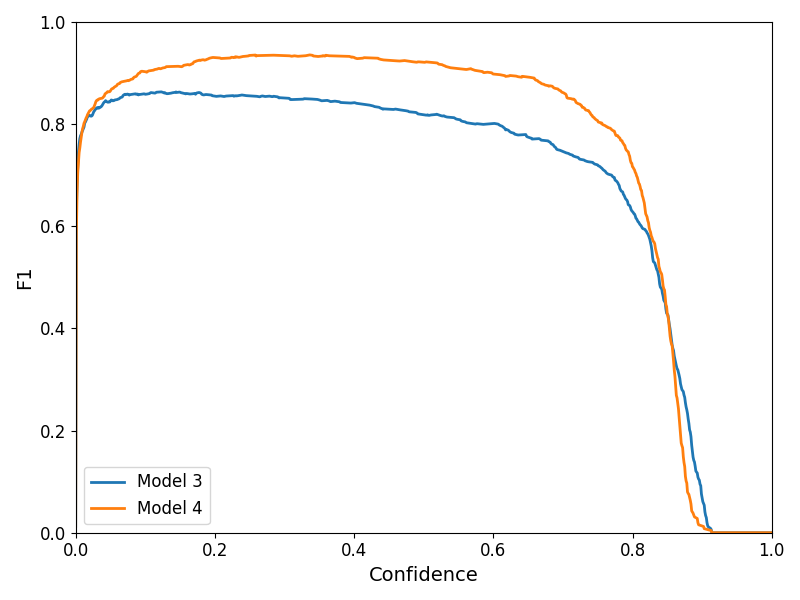} \label{p2z}}
	\caption{Comparison of the AP, PR, and F1 curves before and after adding the small object detection layer.}
	\label{fig:p2}
\end{figure*}
\begin{figure*}[!h]
	\centering
	\subfloat[AP@50]{\includegraphics[width=3.3in]{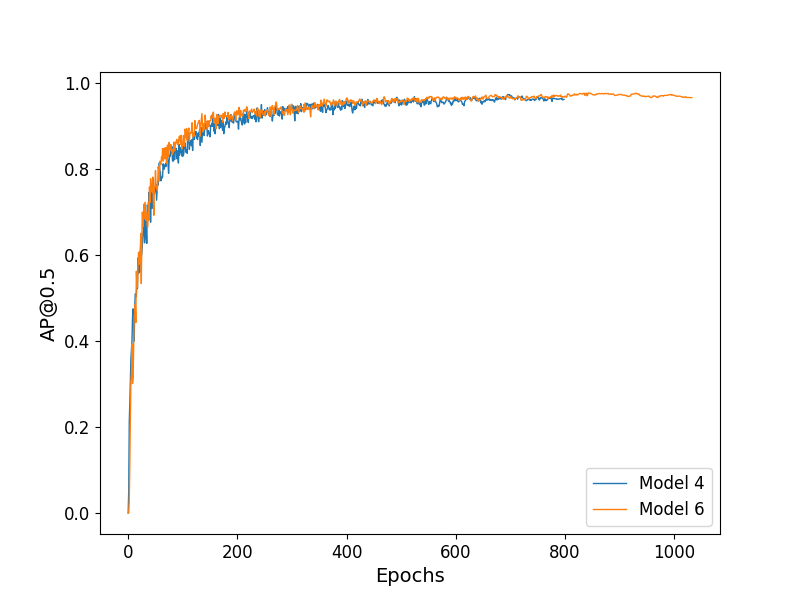} \label{neckw}}
	\hfill
	\subfloat[AP@50-95]{\includegraphics[width=3.3in]{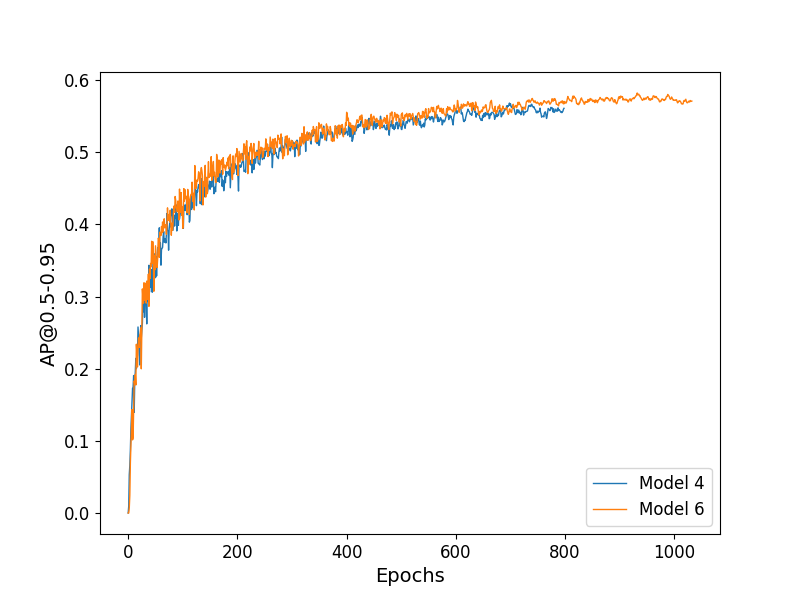} \label{neckx}}
	\vfill
	\subfloat[PR]{\includegraphics[width=3.3in]{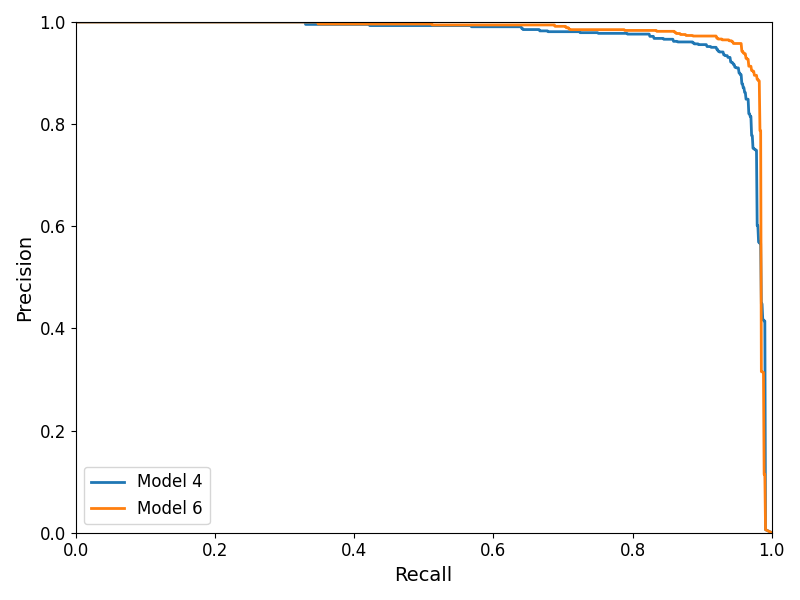} \label{necky}}
	\hfill
	\subfloat[F1]{\includegraphics[width=3.3in]{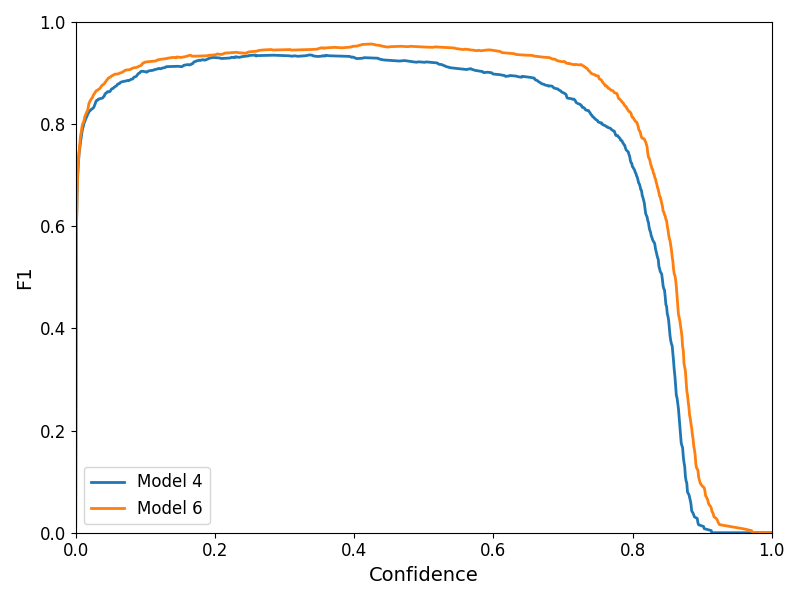} \label{neckz}}
	\caption{Comparison of AP, PR, and F1 curves before and after optimizing the neck.}
	\label{fig:neck}
\end{figure*}
This paper compares our proposed FN-YOLO model with several advanced object detection algorithms, namely YOLOv8, YOLOv9, and YOLOv10, which are among the most advanced detection algorithms currently available.The experiments were conducted on a test set of 250 images with a resolution of \(640 \times 640 \times 3\), and the results are shown in \hyperref[tab:experimental]{Table 1}.

The experimental results presented in \hyperref[tab:algorithm_performance]{Table 2} demonstrate that the proposed FN-YOLO exhibits significant advantages over all other detection algorithms.Specifically, FN-YOLO achieves the lowest parameter count (only \(2.87M\)) and the most compact model size (only \(6.2M\)). Additionally, FN-YOLO excels in processing speed, achieving a rate of \(36.0\) FPS, which is significantly higher than most of the compared algorithms. Most importantly, FN-YOLO substantially outperforms other detection algorithms in terms of precision (\(95.7\%\)), recall (\(94.5\%\)), and average precision (\(97.5\%\)).
Compared to the advanced YOLOv8m, the FN-YOLO model is reduced to \(1/8\) of YOLOv8m's size and the parameter count is reduced to \(1/9\) of YOLOv8m's. Furthermore, FN-YOLO improves precision (\(P\)), recall (\(R\)), and average precision (\(AP\)) by \(0.53\%\), \(9.00\%\), and \(3.50\%\), respectively.

We visualized the inference results of the proposed FN-YOLO model and compared them with those of the YOLOv8, YOLOv9, and YOLOv10 series. The visualization results are shown in \hyperref[fig:reaultvisual]{Figure 9}. It can be observed that YOLOv8, YOLOv9, and YOLOv10 all exhibited instances of missed detections, with the YOLOv9 model missing two dead fish. In contrast, our proposed FN-YOLO accurately detected all dead fish, demonstrating superior performance in the dead fish detection task. Notably, FN-YOLO showed outstanding performance in handling small targets and occlusions, significantly outperforming the other models. This indicates that FN-YOLO possesses higher robustness and accuracy when detecting targets in complex scenes, validating the effectiveness and necessity of our model improvements.
\hyperref[fig:fig10]{Figures 10} and \hyperref[fig:fig11]{Figures 11} show the AP (Average Precision) and PR (Precision-Recall) curves for different models.\hyperref[fig:qipao]{Figure 12} shows a scatter plot comparing the precision and recall of different models. It can be observed that the FN-YOLO model proposed in this paper performs excellently across various evaluation metrics.\hyperref[fig:ld]{Figures 13} presents a performance comparison between YOLOv8s, YOLOv10s, and FN-YOLO. It is evident that FN-YOLO surpasses both YOLOv8s and YOLOv10s in every evaluated aspect, demonstrating comprehensive improvements in precision, recall, AP, parameters, and model size. Specifically, FN-YOLO not only achieves higher values on the AP curve, indicating better overall detection accuracy, but also excels on the PR curve, demonstrating high precision across different recall rates.

Based on the experimental results presented in \hyperref[tab:algorithm_performance]{Table 2} of YOLOv10n, YOLOv10s, and YOLOv10m, we observe that simply increasing the number of channels and layers does not significantly improve the average precision of dead fish detection. Instead, it substantially increases the model parameters and size. This further underscores the effectiveness of our proposed method.

\subsection{Ablation experiment}
To explore the effects of various network modules and enhancements, pertinent ablation experiments were performed. The outcomes of these experiments are compared in \hyperref[tab:ablation experiment]{Table 3}.

In this ablation study, we systematically evaluated the impact of various modules on the performance of the FN-YOLO model for dead fish detection. The baseline model (replacing the YOLOv10 backbone with FasterNet) has a parameter count of 2.02M and serves as a performance benchmark with a precision of 86.5\%, recall of 79.6\%, AP$_{50}$ of 88.0\%, and AP$_{50-95}$ of 47.2\%. Incorporating the SPPF module increased precision to 92.4\% and AP$_{50-95}$ to 51.8\%, although recall slightly decreased to 77.8\%, with the parameter count rising to 2.15M. Further addition of the PSA module improved precision and recall to 89.6\% and 82.6\%, respectively, and AP$_{50-95}$ to 53.7\%, with the parameter count increasing to 2.40M. The inclusion of an additional detection layer significantly enhanced the model’s performance, achieving a precision of 93.4\%, recall of 93.5\%, and AP$_{50-95}$ of 59.8\%, with the parameter count rising to 3.12M.Comparison of the model's performance before and after adding the small object detection layer is shown in \hyperref[fig:p2]{Figure 14}. Incorporating BiFPN path connections reduced the parameter count to 2.89M while further improving precision and recall to 94.4\% and 92.8\%, respectively, and AP$_{50-95}$ to 59.9\%. Replacing the C2f module with the CSPStage module yielded the best performance, with a precision of 95.7\%, recall of 94.5\%, AP$_{50}$ of 97.5\%, and AP$_{50-95}$ of 60.6\%, and a parameter count of 2.87M.\hyperref[fig:neck]{Figure 15} shows the performance comparison of the model before and after optimizing the neck. By comparing experiments 6 and 7, we found that using the traditional Concat method for feature fusion in the Neck part outperformed the weighted feature fusion of BiFPN. In summary, the gradual introduction of these modules significantly improved the model's detection performance, validating the effectiveness and necessity of these enhancements for model optimization.

\begin{table*}[h]
	\centering
	\caption{Dead fish detection ablation experiment results.(1)The SPPF module;(2)The PSA module;(3)Additional detection layer; (4)The path connections in BiFPN ;(5)The CSPStage module(The CSPStage module replaces the C2f module in the neck network);(6)The BiFPN.}
	\label{tab:ablation experiment}
	\begin{tabular}{lccccc}
		\toprule
		Model & Precision (\%) & Recall (\%) & AP$_{50}$ (\%) & AP$_{50-95}$ (\%) & Params (M) \\
		\midrule
		Model 1.FasterNet & 86.5 & 79.6 & 88.0 & 47.2 & \textbf{2.02} \\
		Model 2.FasterNet+(1) & 92.4 & 77.8 & 89.6 & 51.8 & 2.15 \\
		Model 3.FasterNet+(1)+(2) & 89.6 & 82.6 & 91.0 & 53.7 & 2.40 \\
		Model 4.FasterNet+(1)+(2)+(3) & 93.4 & 93.5 & 96.5 & 59.8 & 3.12 \\
		Model 5.FasterNet+(1)+(2)+(3)+(4) & 94.4 & 92.8 & 96.8 & 59.9 & 2.89 \\
		Model 6.FasterNet+(1)+(2)+(3)+(4)+(5) & \textbf{95.7} & \textbf{94.5} & \textbf{97.5} & \textbf{60.6} & 2.87 \\
		Model 7.FasterNet+(1)+(2)+(3)+(5)+(6) & 91.3 & 92.0 & 96.0 & 56.3 & 2.29 \\
		\bottomrule
	\end{tabular}
\end{table*}
\section{Conclusions}
Building upon the YOLOv10 framework, this paper introduces the FN-YOLO model for detecting dead fish on large water surfaces. This model addresses challenges such as small target size, water surface reflections, wave interference, partial submersion, and occlusions from debris. The FasterNet backbone network maintains high feature extraction capabilities while ensuring the model's lightweight nature. Enhancements in feature fusion methods and the incorporation of the CSPStage module significantly improve feature integration efficiency. Additionally, a small target detection layer enhances the model's ability to detect small objects. Experimental results indicate that, compared to the original YOLOv10n model, the P, R, and AP metrics increased by 7.2\%, 10.9\%, and 5.6\%, respectively. Furthermore, compared to models such as YOLOv10m and YOLOv8m, the parameter count was reduced by more than half, while accuracy improved.

These findings demonstrate the proposed method's effectiveness in real-time detection of dead fish on water surfaces. Moreover, due to its low parameter count and computational requirements, the model operates efficiently on low-performance computing devices. This characteristic makes it suitable for deployment in embedded systems, mobile devices, and other resource-constrained environments, facilitating its practical implementation in production settings.

\section*{CRediT authorship contribution statement }
\textbf{Qingbin Tian:} Conceptualization, Methodology, Software, Investigation, Writing – original draft, Visualization, Writing – review \& editing.\textbf{Yukang Huo:} Data curation, Validation, Investigation.\textbf{Mingyuan Yao:} Data curation, Investigation.\textbf{Yugang Cai:} Resources, Project administration.
\textbf{Haihua Wang:} Conceptualization, Supervision, Funding acquisition, Writing – review \& editing.

\section*{Declaration of Competing Interest}

The authors declare that they have no known competing financial 
interests or personal relationships that could have appeared to influence the work reported in this paper. 
\section*{Acknowledgments}









\bibliographystyle{cas-model2-names}

\bibliography{cas-refs}



\end{document}